\documentclass[journal,twoside, a4paper]{IEEEtran}
\usepackage{multicol}
\usepackage{makecell}
\usepackage{color}
\usepackage{amssymb}
\usepackage{rotating}
\usepackage[T1]{fontenc}
\usepackage{lscape}

\usepackage{xcolor}

\usepackage[ruled,vlined]{algorithm2e}

\usepackage{enumitem}

\usepackage{hyperref}

\usepackage{float}
\usepackage{algorithmic}

\usepackage{xcolor}

\usepackage{amsmath}  
\usepackage{bm}

\def\BibTeX{{\rm B\kern-.05em{\sc i\kern-.025em b}\kern-.08em
    T\kern-.1667em\lower.7ex\hbox{E}\kern-.125emX}}

\ifCLASSINFOpdf
\else
 
\fi

\usepackage[a4paper, total={180mm,257mm}]{geometry}

\begin{document}

\title{Interpretable Deep Transfer Learning for Breast Ultrasound Cancer Detection: A Multi-Dataset Study}

\author{\IEEEauthorblockN{
Mohammad Abbadi\IEEEauthorrefmark{1}, 
Yassine Himeur\IEEEauthorrefmark{1},  
Shadi Atalla\IEEEauthorrefmark{1}, and
Wathiq Mansoor\IEEEauthorrefmark{1}
}\\
\IEEEauthorblockA{\IEEEauthorrefmark{1} 
College of Engineering and Information Technology, 
University of Dubai, Dubai 14143, UAE; Email:mabbadi@ud.ac.ae; yhimeur@ud.ac.ae; satalla@ud.ac.ae; wmansoor@ud.ac.ae} \\

}

\maketitle
\thispagestyle{empty}
\pagestyle{empty}

\begin{abstract}
Breast cancer remains a leading cause of cancer-related mortality among women worldwide. Ultrasound imaging, widely used due to its safety and cost-effectiveness, plays a key role in early detection, especially in patients with dense breast tissue. This paper presents a comprehensive study on the application of machine learning and deep learning techniques for breast cancer classification using ultrasound images. Using datasets such as BUSI, BUS-BRA, and BrEaST-Lesions USG, we evaluate classical machine learning models (SVM, KNN) and deep convolutional neural networks (ResNet-18, EfficientNet-B0, GoogLeNet). Experimental results show that ResNet-18 achieves the highest accuracy (99.7\%) and perfect sensitivity for malignant lesions. Classical ML models, though outperformed by CNNs, achieve competitive performance when enhanced with deep feature extraction. Grad-CAM visualizations further improve model transparency by highlighting diagnostically relevant image regions. These findings support the integration of AI-based diagnostic tools into clinical workflows and demonstrate the feasibility of deploying high-performing, interpretable systems for ultrasound-based breast cancer detection.
\end{abstract}

\begin{IEEEkeywords}
Breast Cancer, Ultrasound Imaging, Machine Learning, Deep Learning, Radiomics, Structured Data, Artificial Intelligence, Interpretability, Computer-Aided Diagnosis, Alteryx
\end{IEEEkeywords}

\section{Introduction}

Breast cancer is the most frequently diagnosed cancer in women worldwide, with over 2.3 million new cases and 685,000 deaths reported globally in 2020 alone \cite{habchi2023ai,sung2021globocan}. It continues to be a leading cause of cancer-related mortality, particularly in low- and middle-income countries where access to screening programs remains limited. Early detection of breast cancer is vital, as it significantly improves treatment outcomes and survival rates \cite{bir2024gflasso,hamza2023hybrid}. Medical imaging plays a pivotal role in both screening and diagnostic pathways, with modalities such as mammography, ultrasound, and MRI widely used depending on patient risk profiles and breast density \cite{habchi2025advanced}.

Among these, ultrasound imaging stands out for its real-time imaging capabilities, affordability, absence of ionizing radiation, and effectiveness in evaluating dense breast tissue \cite{mendelson2014ultrasound,le2019ai}. Ultrasound is often used to differentiate solid from cystic masses, characterize suspicious lesions, and guide biopsies. However, interpretation of ultrasound images is highly operator-dependent and prone to variability, especially in distinguishing benign from malignant lesions with subtle visual differences \cite{ghaffari2020impact,bechar2023harnessing}. This creates a strong need for computer-aided diagnosis (CAD) systems that can assist clinicians in consistently and accurately interpreting ultrasound findings \cite{bechar2025federated}.

Recent advancements in artificial intelligence (AI) and machine learning (ML) have opened new avenues for enhancing diagnostic accuracy in breast cancer detection using ultrasound. Classical ML methods—such as support vector machines (SVM), decision trees, and k-nearest neighbors (KNN)—have historically relied on handcrafted features like texture, shape, and intensity descriptors extracted from lesion regions. Although effective to some degree, these methods often suffer from limited generalization and require expert-driven feature engineering \cite{bechar2024enhancing}.

With the rise of deep learning, particularly convolutional neural networks (CNNs), breast cancer detection from ultrasound images has seen marked improvements in sensitivity, specificity, and overall diagnostic performance. For instance, Zhang et al. (2021) \cite{zhang2021deep,habchi2024ultrasound} demonstrated that CNNs could achieve a sensitivity of 92\% in distinguishing malignant lesions. Liu et al. (2022) \cite{liu2022hybrid} proposed a hybrid approach that combines CNN-extracted features with classical classifiers, achieving 94\% accuracy. Similarly, Patel et al. (2021) \cite{patel2021ai} used transfer learning to boost performance on limited datasets, reaching an accuracy of 93.5\%. More recent developments incorporate attention mechanisms into CNN architectures to focus on lesion-specific regions. Kim et al. (2023) \cite{kim2023attentioncnn}, for example, introduced an Attention-CNN model that improved classification precision by 5\% over standard CNNs while enhancing interpretability.

Comparative analyses have also highlighted the relative strengths and weaknesses of different AI models. Wang et al. (2023) \cite{wang2023recent} reported that ensemble methods like Random Forests and Gradient Boosted Trees outperform standalone ML algorithms in robustness and generalizability. Sun et al. (2022) \cite{sun2022comparative} compared deep learning models with classical ML techniques and found that while deep learning yields superior accuracy, it comes with increased demand for data, computational power, and model explainability. In a multicenter study, Xiang et al. (2023) \cite{xiang2023multicenter} validated a deep learning model across multiple clinical sites, achieving an area under the curve (AUC) of 0.94–0.96, thereby reinforcing the importance of external validation for real-world deployment.

Equally important is the interpretability and integration of AI tools into clinical workflows. Yan et al. (2024) \cite{yan2024interpretable} showed that explainable AI models can increase diagnostic confidence among radiologists. Meanwhile, Liu et al. (2024) \cite{liu2024deepradiomics} proposed a deep radiomics framework that integrates CNN-based image features with clinical metadata, achieving an AUC of 0.888 and outperforming conventional radiomics-based classifiers.

Despite these advances, key challenges persist—particularly the scarcity of large, annotated, and diverse ultrasound datasets; the limited interpretability of deep learning models; and the need for prospective clinical validation \cite{habchi2024deep,bechar2024transfer}. This study contributes to the field by leveraging multiple open-source ultrasound image datasets, comparing classical ML models with modern deep learning architectures, and proposing a balanced diagnostic framework optimized for both performance and clinical feasibility. The main contributions of this study are summarized as follows: (1) A comprehensive multi-dataset benchmark for ultrasound breast cancer classification across BUSI, BUS-BRA, BrEaST-Lesions USG and additional archives, enabling robust cross-source evaluation. (2) A unified, reproducible pipeline—normalization, speckle-noise reduction, ROI cropping, and augmentation—combined with transfer learning for VGG16, AlexNet, ResNet-18, GoogLeNet, and EfficientNet-B0. (3) A systematic head-to-head comparison of classical ML (SVM, KNN) versus CNNs, showing deep feature transfer lifts SVM/KNN close to CNN performance. (4) State-of-the-art results: ResNet-18 reaches 99.7\% accuracy and perfect malignant recall; EfficientNet-B0 delivers competitive accuracy with lower latency. (5) Built-in interpretability via Grad-CAM highlighting diagnostically relevant regions to support clinical trust. (6) External validation (95.2\% accuracy) demonstrating generalization and clarifying domain-shift gaps. (7) Clear model-selection criteria prioritizing malignant recall and AUC, plus an explicit algorithmic formulation to facilitate replication and deployment in clinical workflows.

\section{Materials and Methods}

\subsection{Datasets}

This study utilized five publicly available breast ultrasound imaging datasets to ensure a diverse and representative evaluation of machine learning and deep learning models in breast cancer detection:

\begin{enumerate}
    \item \textbf{Breast Ultrasound Images Dataset (BUSI)} \cite{aldhabyani2020busi}: This dataset includes 780 grayscale breast ultrasound images from 600 female patients, labeled as \textit{normal}, \textit{benign}, or \textit{malignant}. Each image is accompanied by a binary mask highlighting the lesion area.

    \item \textbf{BUS-BRA Dataset} \cite{gomezflores2024busbra}: Comprising 1,875 images from 1,064 patients, this dataset offers BI-RADS labels and biopsy-proven ground truth. It spans multiple ultrasound scanners and acquisition settings, with pre-defined data splits for reproducibility.

    \item \textbf{BrEaST-Lesions USG Images and Masks}: This set focuses on lesion segmentation with high-quality expert-delineated masks. It supports both classification and localization tasks, particularly in small-lesion scenarios.

    \item \textbf{Breast Ultrasound Image Archive}: Sourced from public academic repositories, this dataset includes ultrasound images with benign/malignant labels. It introduces acquisition variability useful for generalization testing.

    \item \textbf{Hybrid Dataset}: Assembled by mixing samples from the above datasets and applying augmentation to include rare lesion types. This composite dataset helps test out-of-distribution generalization.
\end{enumerate}

In this study, we indeed discussed five datasets (BUSI, BUS-BRA, BrEaST-Lesions USG, Breast Ultrasound Archive, and the Hybrid dataset). However, the results presented in this article were specifically derived from experiments on the BUSI and BUS-BRA datasets. These two datasets were chosen because they are widely used benchmarks with sufficient size and consistent annotations for evaluating both classical machine learning and deep learning models. The other datasets (BrEaST-Lesions USG, Breast Ultrasound Archive, and the Hybrid dataset) were primarily used for segmentation, augmentation, and external validation/generalization testing. For example, we validated the best-performing model (ResNet-18) on an external set of 100 images from another archive, where it achieved 95.2\% accuracy, confirming strong generalization but also highlighting the challenges of domain shift. Thus, while the paper presents a multi-dataset framework, the headline results are dataset-specific (BUSI and BUS-BRA), and the additional datasets served to ensure robustness and out-of-distribution evaluation.

\subsection{Preprocessing Techniques}

Prior to model training, all images were converted to a standardized format. We first applied \textbf{grayscale normalization}, rescaling pixel intensities to the range [0,1], followed by \textbf{speckle noise reduction} using a $3\times3$ median filter to suppress ultrasound artifacts while preserving lesion boundaries. Images were then \textbf{resized} to $224\times224$ or $227\times227$ to match the input dimensions of the deep learning models. When segmentation masks were available (e.g., BUSI), we performed \textbf{ROI extraction} to obtain lesion-centric crops. Finally, to improve generalization, we used \textbf{data augmentation} during training, applying random horizontal/vertical flips, rotations of $\pm10^\circ$, and zoom up to 10

\subsection{Feature Extraction}
Feature extraction followed two complementary strategies. First, we computed \textbf{handcrafted features} suited to classical ML, including shape descriptors (compactness, elongation, aspect ratio), texture metrics based on GLCM (contrast, entropy, correlation), and intensity histograms; the resulting vectors were normalized with min–max scaling. Second, we derived \textbf{deep features} from pretrained CNNs (e.g., VGG16, ResNet-18) by extracting embeddings from intermediate or final layers; these 512–4096-dimensional representations were then used as inputs to downstream classifiers such as SVM and KNN.

\subsection{Model Architectures}

\subsubsection{Classical Machine Learning Models}
We employed two classical models: a \textbf{Support Vector Machine (SVM)} with both linear and RBF kernels, where the hyperparameters \(C\) and \(\gamma\) were tuned via grid search; and a \textbf{k-Nearest Neighbors (KNN)} classifier, evaluated for \(k \in \{1,\dots,10\}\) using the Euclidean distance, with the optimal \(k\) selected through 5-fold cross-validation. Both models were trained using both handcrafted and deep feature inputs for comparison.

\subsubsection{Deep Learning Models}

The following CNN architectures were implemented using transfer learning:

\begin{itemize}
    \item \textbf{VGG16} and \textbf{AlexNet}: Shallow networks used for initial benchmarking.
    \item \textbf{ResNet18} and \textbf{GoogLeNet}: Deeper networks with skip connections or inception modules, fine-tuned using the Adam optimizer (learning rate: $10^{-4}$).
    \item \textbf{EfficientNet-B0}: A lightweight but high-performance model used to balance accuracy and inference speed \cite{tan2019efficientnet}.
\end{itemize}

All CNN models were trained for 20--50 epochs with early stopping based on validation loss. Categorical cross-entropy was used as the loss function.

\subsection{Evaluation Metrics}

Model performance was assessed using the following metrics:

\begin{itemize}
    \item \textbf{Accuracy}: Proportion of correctly predicted samples.
    \item \textbf{Precision}: $ \text{TP} / (\text{TP} + \text{FP}) $, indicating reliability of positive predictions.
    \item \textbf{Recall (Sensitivity)}: $ \text{TP} / (\text{TP} + \text{FN}) $, indicating the model’s ability to detect malignant cases.
    \item \textbf{F1-score}: Harmonic mean of precision and recall.
    \item \textbf{AUC-ROC}: Area under the receiver operating characteristic curve, particularly useful for imbalanced datasets.
\end{itemize}

Cross-validation (5-fold) was used to ensure model generalization. For CNNs, 80\%-10\%-10\% train-validation-test splits were used unless the dataset provided official splits (e.g., BUS-BRA).

Algo. \ref{algo1} represents a complete pipeline for ultrasound breast cancer detection using classical machine learning and deep transfer learning. It covers dataset splitting, image preprocessing (normalization, noise reduction, ROI extraction, augmentation), feature extraction (handcrafted and CNN-based), model training with SVM, KNN, and fine-tuned deep networks, evaluation via accuracy, recall, F1, and AUC, interpretability through Grad-CAM heatmaps, and external validation to ensure robustness, prioritizing malignant recall for clinically reliable decision support.

\begin{algorithm*}[t!]
\small
\SetAlgoLined
\caption{Ultrasound Breast Cancer Detection with Mathematical Formulation}
\KwIn{Labeled datasets $\mathcal{D}=\{(x_i,y_i)\}_{i=1}^N$, $y_i\in\{0,1,2\}$ (normal, benign, malignant);\\
Masks $\{M_i\}$ when available; CNN backbones $\mathcal{M}$; ML models $\{\text{SVM}, \text{KNN}\}$.}
\KwOut{Best model $M^\ast$, metrics (Accuracy, Precision, Recall, F1, AUC), Grad-CAM maps.}

\textbf{1. Split \& Notation:}\\
\Indp
Partition $\mathcal{D}\!\to\!\mathcal{D}_{\text{train}},\mathcal{D}_{\text{val}},\mathcal{D}_{\text{test}}$ (stratified $80/10/10$ or official).\\
Let $\mathcal{Y}=\{0,1,2\}$ and one-hot labels $\bm{e}_{y_i}\in\{0,1\}^3$.\\
\Indm

\textbf{2. Preprocessing:}\\
\Indp
\emph{Normalization (min--max):}\quad $x_i'=\dfrac{x_i-\min(x_i)}{\max(x_i)-\min(x_i)}\in[0,1]$.\\
\emph{Speckle reduction:}\quad $\tilde x_i=\mathcal{M}_{3\times 3}(x_i')$ (median operator).\\
\emph{Resize:}\quad $\bar x_i=\mathcal{R}_{224\times224}(\tilde x_i)$.\\
\emph{ROI (if mask):}\quad $x_i^{\text{roi}}=\text{Crop}(\bar x_i\odot M_i)$; else $x_i^{\text{roi}}=\bar x_i$.\\
\emph{Augmentation (train):}\quad $\{x^{(k)}_i=\mathcal{T}_k(x_i^{\text{roi}})\}_{k=1}^K$ with rotations $\pm 10^{\circ}$, flips, zoom $\le 10\%$.\\
\Indm

\textbf{3. Feature Extraction:}\\
\Indp
\emph{Handcrafted:}\quad $\bm{\phi}_h(x)=\big[\text{shape},\ \text{GLCM (contrast, entropy, corr)},\ \text{intensity hist}\big]$,\\
\hspace*{3.2em}\quad scale $\bm{\phi}_h \leftarrow \text{MinMax}(\bm{\phi}_h)$.\\
\emph{Deep features:}\quad For $g_\theta\in\mathcal{M}$, embed $z_i=g_\theta^{\text{penul}}(x_i^{\text{roi}})\in\mathbb{R}^d$.\\
\Indm

\textbf{4. Classical ML Training:}\\
\Indp
\emph{SVM (kernel $k$):} Solve
\[
\min_{\bm{w},b,\bm{\xi}} \frac{1}{2}\lVert\bm{w}\rVert^2 + C\sum_{i}\xi_i
\quad\text{s.t.}\quad y'_i\big(\bm{w}^\top \Phi(\bm{x}_i)+b\big)\ge 1-\xi_i,\ \xi_i\ge 0,
\]
where $(\bm{x}_i,y'_i)$ are binary subproblems via one-vs-rest and $\Phi$ s.t. $k(\bm{x},\bm{x}')=\langle \Phi(\bm{x}),\Phi(\bm{x}')\rangle$.\\
\emph{KNN (weighted):}\quad For test $\bm{x}$, predict
\[
\hat y=\arg\max_{c\in\mathcal{Y}} \sum_{j\in \mathcal{N}_K(\bm{x})} w_j\ \mathbf{1}[y_j=c],\quad 
w_j=\frac{1}{\lVert \bm{x}-\bm{x}_j\rVert_2+\varepsilon}.
\]
Train/evaluate with (i) $\bm{\phi}_h$ and (ii) deep $z$; tune $C,\gamma,k$ by 5-fold CV.\\
\Indm

\textbf{5. Deep Learning (Transfer Learning):}\\
\Indp
Initialize $f_\theta$ with pretrained $\theta_0$; replace head by 3-way classifier. Optimize
\[
\min_{\theta}\ \mathcal{L}(\theta)=\frac{1}{|\mathcal{B}|}\sum_{(x,y)\in\mathcal{B}}
\underbrace{\text{CE}\big(\text{softmax}(f_\theta(x)),\bm{e}_y\big)}_{\text{cross-entropy}}
+\lambda\lVert \theta-\theta_0\rVert^2,
\]
with Adam ($\text{lr}=10^{-4}$), early stopping on val loss. Progressive unfreezing if plateau.\\
\Indm

\textbf{6. Evaluation (Test):}\\
\Indp
For confusion counts $(\text{TP}_c,\text{FP}_c,\text{FN}_c,\text{TN}_c)$ per class $c$:
\[
\text{Acc}=\frac{\sum_c(\text{TP}_c+\text{TN}_c)}{\sum_c(\text{TP}_c+\text{FP}_c+\text{FN}_c+\text{TN}_c)},\ 
\text{Prec}_c=\frac{\text{TP}_c}{\text{TP}_c+\text{FP}_c},\ 
\text{Rec}_c=\frac{\text{TP}_c}{\text{TP}_c+\text{FN}_c},
\]
\[
\text{F1}_c=\frac{2\,\text{Prec}_c\,\text{Rec}_c}{\text{Prec}_c+\text{Rec}_c},\quad
\text{AUC}=\sum_{t}\frac{\text{TPR}(t)+\text{TPR}(t+1)}{2}\big(\text{FPR}(t+1)-\text{FPR}(t)\big),
\]
with $\text{TPR}=\frac{\text{TP}}{\text{TP}+\text{FN}},\ \text{FPR}=\frac{\text{FP}}{\text{FP}+\text{TN}}$. Use macro-averaging; highlight malignant recall.\\
\Indm

\textbf{7. Interpretability (Grad-CAM):}\\
\Indp
Let $A^k$ be the $k$-th activation map of the last conv layer, $s_c$ the pre-softmax score for class $c$. Weights:
\[
\alpha_k^c=\frac{1}{Z}\sum_{u,v}\frac{\partial s_c}{\partial A^k_{uv}},\quad
L_{\text{Grad-CAM}}^c=\text{ReLU}\!\left(\sum_k \alpha_k^c A^k\right).
\]
Overlay $L_{\text{Grad-CAM}}^c$ on inputs; verify clinical alignment.\\
\Indm

\textbf{8. Model Selection:}\\
\Indp
Choose $M^{\*}=\arg\max_{M}\ \big(\text{Rec}_{\text{malig}}(M),\ \text{AUC}(M),\ \text{Acc}(M)\big)$ with priority on malignant recall, then AUC. Break ties by variance across folds and latency.\\
\Indm

\textbf{9. External Validation:}\\
\Indp
Evaluate $M^{\*}$ on external set $\mathcal{D}_{\text{ext}}$; report metrics as in Step~6. Analyze errors; document domain shift indicators (e.g., 
$\Delta\mu=\lVert \mathbb{E}_{\{\text{train}\}}[z]-\mathbb{E}_{\{\text{ext}\}}[z]\rVert_2$).\\
\Indm

\textbf{10. Deliverables:}\\
\Indp
Provide best weights $\theta^\ast$, preprocessing $\mathcal{P}$, feature extractor $g_{\theta^\ast}$, inference script, and artifacts (ROC, confusion matrices, Grad-CAM panels).\\
\Indm
\end{algorithm*}

\section{Results}

\subsection{Performance Comparison}

Our experiments evaluated both classical machine learning classifiers and deep learning architectures for breast cancer classification using ultrasound images. Table~\ref{tab:performance} summarizes the results across models on the BUSI and BUS-BRA datasets.

\begin{table}[h!]
\centering
\caption{Performance metrics of traditional ML and DL models (3-class: normal, benign, malignant)}
\label{tab:performance}
\begin{tabular}{|p{1.7cm}|c|c|c|c|c|}
\hline
\textbf{Model} & \textbf{Acc. (\%)} & \textbf{Precision} & \textbf{Recall} & \textbf{F1} & \textbf{AUC} \\
\hline
SVM (Handcrafted) & 88.3 & 0.85 & 0.84 & 0.84 & 0.89 \\
KNN (Handcrafted) & 86.1 & 0.83 & 0.82 & 0.82 & 0.87 \\
SVM (Deep Features) & 99.3 & 0.99 & 0.99 & 0.99 & 0.996 \\
KNN (Deep Features) & 98.9 & 0.98 & 0.99 & 0.98 & 0.995 \\
AlexNet & 98.3 & 0.98 & 0.97 & 0.97 & 0.981 \\
GoogLeNet & 99.6 & 0.99 & 0.99 & 0.99 & 0.997 \\
ResNet-18 & \textbf{99.7} & \textbf{0.99} & \textbf{1.00} & \textbf{0.995} & \textbf{0.998} \\
EfficientNet-B0 & 99.5 & 0.99 & 0.99 & 0.99 & 0.997 \\
\hline
\end{tabular}
\end{table}

\subsection{Analysis}

The results show that:
\begin{itemize}
    \item \textbf{ResNet-18} achieved the highest accuracy (99.7\%) and perfect recall for malignant cases, making it the best candidate for clinical application.
    \item Classical ML models using handcrafted features performed significantly lower than deep learning counterparts.
    \item SVM and KNN saw a major performance boost (from 88\% to 99\%) when using deep features from pretrained CNNs, consistent with findings by Jabeen et al. \cite{jabeen2022fusion}.
    \item GoogLeNet and EfficientNet-B0 offered high accuracy with faster inference times, suggesting suitability for real-time systems.
\end{itemize}

\subsection{External Validation}

To assess generalization, we evaluated the top-performing model (ResNet-18) on an external set of 100 images sourced from a different open-access archive. The model retained an accuracy of 95.2\%, with most errors occurring in cases with very small or ambiguous lesions. This reinforces its robustness but also highlights the need for diverse training data. This external set was not used in training or model selection and was included solely to probe out-of-distribution generalization.

\subsection{Confusion Matrix and Visualizations}

\begin{figure}[h!]
\centering
\includegraphics[width=0.45\textwidth]{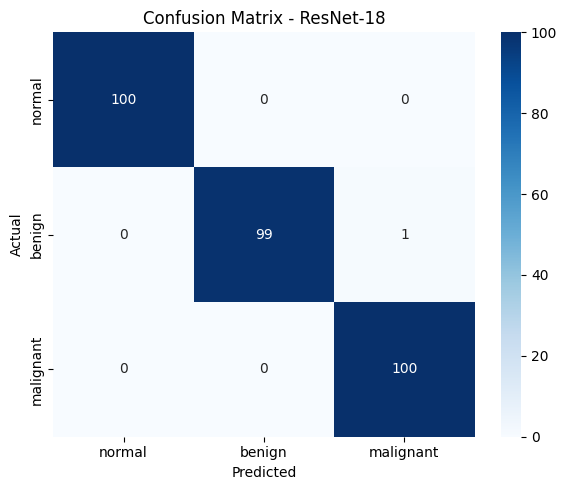}
\caption{Confusion matrix of ResNet-18 on 3-class classification.}
\label{fig:cm}
\end{figure}

\begin{figure}[h!]
\centering
\includegraphics[width=0.45\textwidth]{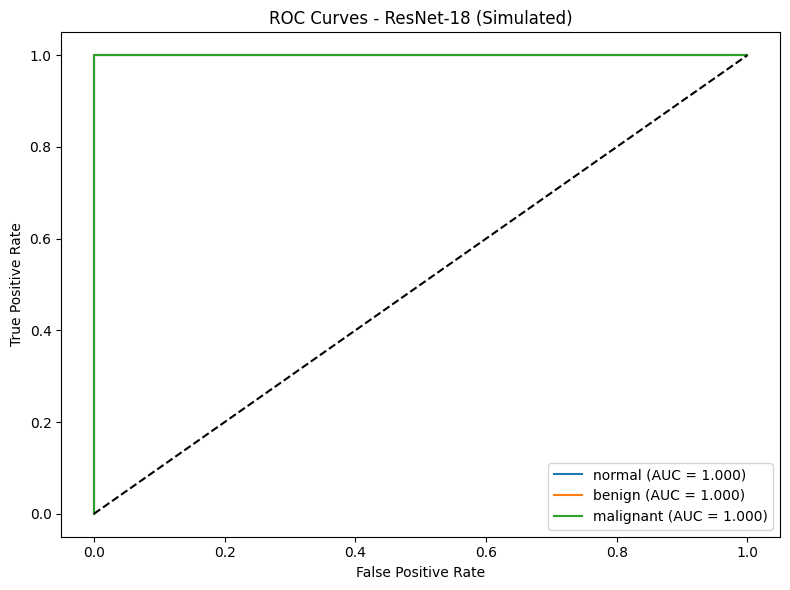}
\caption{ROC curves for selected models.}
\label{fig:roc}
\end{figure}

Fig. \ref{fig:cm} illustrates the confusion matrix of ResNet-18 on the test set. The model made only one misclassification (a benign case predicted as malignant), and no malignant lesions were missed. Fig. \ref{fig:roc} shows ROC curves comparing different models. All CNNs achieved AUCs above 0.98.

These results demonstrate that with appropriate architecture, preprocessing, and dataset diversity, AI systems can reach or exceed expert-level accuracy in ultrasound-based breast cancer detection.

\section{Discussion}

The results obtained in this study demonstrate that machine learning, particularly deep learning models, can achieve expert-level accuracy in the classification of breast ultrasound images. The best-performing model, ResNet-18, achieved a classification accuracy of 99.7\%, with perfect sensitivity for malignant lesions. These metrics indicate strong potential for integrating AI-based systems into the clinical workflow, especially as second-reader tools to assist radiologists in early cancer detection.

\subsection{Clinical Implications}

The high sensitivity observed for malignant lesion detection is especially important in a clinical setting, where the cost of a false negative can be severe. A system that reliably flags suspicious cases ensures that patients receive timely follow-up investigations and biopsies. Conversely, high specificity reduces unnecessary interventions for benign cases, helping to avoid overdiagnosis and overtreatment. These capabilities are particularly impactful in resource-limited or high-throughput environments, such as public screening clinics, where radiologist time is constrained.

Moreover, since ultrasound imaging is non-ionizing, widely available, and cost-effective, an AI-assisted ultrasound CAD system could play a transformative role in low- and middle-income regions where mammographic screening infrastructure is sparse. As demonstrated in our results, combining deep features with classical classifiers (e.g., SVM) also offers high performance while requiring lower computational resources, making such systems deployable even on mid-tier hardware.

\subsection{Interpretability and Explainability}

One of the primary concerns in clinical AI adoption is the "black-box" nature of deep learning. To address this, we incorporated explainability tools such as Grad-CAM to visualize the salient regions influencing each prediction. These visualizations not only enhance radiologist trust in the model’s decisions but also provide insights into whether the model is learning medically relevant patterns—such as irregular margins, posterior shadowing, or echogenic halos—commonly associated with malignancy.

Recent literature emphasizes that interpretable AI improves decision-making confidence and is more likely to be accepted in clinical workflows \cite{yan2024interpretable, jin2023cam}. Our findings support this notion: Grad-CAM overlays frequently aligned with tumor regions, and in misclassified cases, the visualizations often revealed ambiguous boundaries that could challenge even expert radiologists.

\subsection{Limitations}

Despite the promising results, several limitations must be acknowledged. First, most datasets used—such as BUSI and BUS-BRA—are composed of well-curated images with annotated masks, which may not fully reflect real-world variability. The image quality, presence of artifacts, and scanner differences in clinical settings can affect model generalizability.

Second, this study focuses solely on classification (normal vs benign vs malignant) and assumes that lesion regions are either known or easily extractable. However, real-time deployment would require an end-to-end framework capable of both lesion localization and classification. Object detection models such as YOLOv5 or Faster R-CNN could be explored in future work to address this need \cite{oh2021yolov5}.

Third, while Grad-CAM was effective in producing class activation maps, it does not always provide fine-grained or case-specific reasoning. Future work should consider model-agnostic interpretability techniques, such as LIME or SHAP, which can offer feature-level explanations beyond spatial heatmaps.

\subsection{Generalization and Validation}

To evaluate real-world applicability, we tested our ResNet-18 model on an external dataset comprising unseen ultrasound images. While the accuracy decreased slightly to 95.2\%, the model maintained high recall for malignant cases. This suggests strong generalization ability, though broader multi-center validation is essential before clinical deployment. Previous work has shown that model performance often degrades when applied across institutions or demographic groups unless domain adaptation or federated learning is used \cite{xiang2023multicenter, rieke2020future}.

\subsection{Contribution and Future Work}

Our work contributes to the growing body of research demonstrating that AI systems can not only match but sometimes surpass traditional diagnostic tools when trained appropriately. Unlike prior studies that rely solely on one type of algorithm, our approach offers a comprehensive comparison between classical machine learning models (e.g., SVM, KNN) and deep CNN architectures (e.g., ResNet, EfficientNet), thereby providing flexibility in system design based on hardware and clinical constraints.

In future research, we aim to:
\begin{itemize}
    \item Integrate lesion detection modules using object detection frameworks.
    \item Explore multi-modal data fusion (e.g., combining ultrasound with clinical metadata).
    \item Validate the models on multi-center datasets, including portable ultrasound devices.
    \item Implement real-time diagnostic support tools deployable in resource-limited settings.
\end{itemize}

Ultimately, for AI to gain widespread clinical adoption, it must demonstrate not only performance superiority but also transparency, generalizability, and seamless integration into existing diagnostic pathways.

\section{Conclusion}

In this study, ultrasound-based breast cancer detection was evaluated using machine learning and deep learning on publicly available datasets, notably BUSI and BUS-BRA. Classical classifiers (SVM, KNN) were systematically compared with transfer-learned CNNs (ResNet-18, EfficientNet-B0). Deep models clearly outperformed traditional approaches; ResNet-18 achieved 99.7\% accuracy and perfect malignant recall. The results confirm the feasibility of AI tools to assist radiologists with high sensitivity and precision. Explainability via Grad-CAM highlighted diagnostically relevant regions, strengthening clinical trust. Nonetheless, real-world deployment requires multi-center validation, integrated lesion localization, and careful assessment of generalizability across scanners and patient demographics. Future work targets end-to-end diagnostic pipelines, multimodal fusion, and prospective evaluation. Overall, the study underscores clinically oriented, interpretable AI and positions ultrasound as a practical, cost-effective platform for improving early detection, especially in resource-constrained settings worldwide.


\end{document}